\documentclass[11pt,a4paper]{article}
\usepackage[hyperref]{emnlp-ijcnlp-2019}
\usepackage{times}
\usepackage{latexsym}
\usepackage{graphicx}
\usepackage{amsmath}
\usepackage{amsfonts}
\usepackage{booktabs}
\usepackage{multirow}
\usepackage{bm}
\usepackage[english]{babel}

\usepackage{url}
\usepackage{array}

\usepackage{caption}
\newcommand{\PreserveBackslash}[1]{\let\temp=\\#1\let\\=\temp}
\newcolumntype{C}[1]{>{\PreserveBackslash\centering}p{#1}}
\newcolumntype{R}[1]{>{\PreserveBackslash\raggedleft}p{#1}}
\newcolumntype{L}[1]{>{\PreserveBackslash\raggedright}p{#1}}

\aclfinalcopy 

\title{Modeling Event Background for If-Then Commonsense Reasoning Using Context-aware Variational Autoencoder}

\author {\textbf{Li Du, Xiao Ding, Ting Liu\thanks{Corresponding author}} \, and \textbf{Zhongyang Li} \\
        Research Center for Social Computing and Information Retrieval \\
        Harbin Institute of Technology, China \\
        \{ldu, xding, tliu, zyli\}@ir.hit.edu.cn
        }
        
\date{}

\begin{document}
\maketitle
\begin{abstract}
 Understanding event and event-centered commonsense reasoning are crucial for natural language processing (NLP). Given an observed event, it is trivial for human to infer its intents and effects, while this type of If-Then reasoning still remains challenging for NLP systems. To facilitate this, a If-Then commonsense reasoning dataset Atomic is proposed, together with an RNN-based Seq2Seq model to conduct such reasoning. However, two fundamental problems still need to be addressed: first, the intents of an event may be multiple, while the generations of RNN-based Seq2Seq models are always semantically close; second, external knowledge of the event background may be necessary for understanding events and conducting the If-Then reasoning. To address these issues, we propose a novel context-aware variational autoencoder effectively learning event background information to guide the If-Then reasoning. Experimental results show that our approach improves the accuracy and diversity of inferences compared with state-of-the-art baseline methods.
\end{abstract}

\newcommand{\tabincell}[2]{\begin{tabular}{@{}#1@{}}#2\end{tabular}}

\section{Introduction}
Recently, event-centered commonsense knowledge has attracted much attention \cite{chambers2008unsupervised,SEGERS16.722,wang2017integrating,li2018constructing}, because of understanding events is an important component of NLP. Given a daily-life event, human can easily understand it and reason about its causes, effects, and so on. However, it still remains a challenging task for NLP
systems. This is partly due to most of them are trained for
task-specific datasets or objectives, which results in models
that are adapt at finding task-specific underlying correlation
patterns but have limited capability in simple and explainable
commonsense reasoning \cite{sap2018atomic}.

To facilitate this, \citeauthor{rashkin2018event2mind} (\citeyear{rashkin2018event2mind}) build the Event2Mind dataset and \citeauthor{sap2018atomic} (\citeyear{sap2018atomic}) present the Atomic dataset, mainly focus on nine If-Then reasoning types to describe causes, effects, intents and participant characteristic about events. Together with these datasets, a simple RNN-based encoder-decoder framework is proposed to conduct the If-Then reasoning.

\begin{figure}
  \centering
  \includegraphics[width=1.001\linewidth]{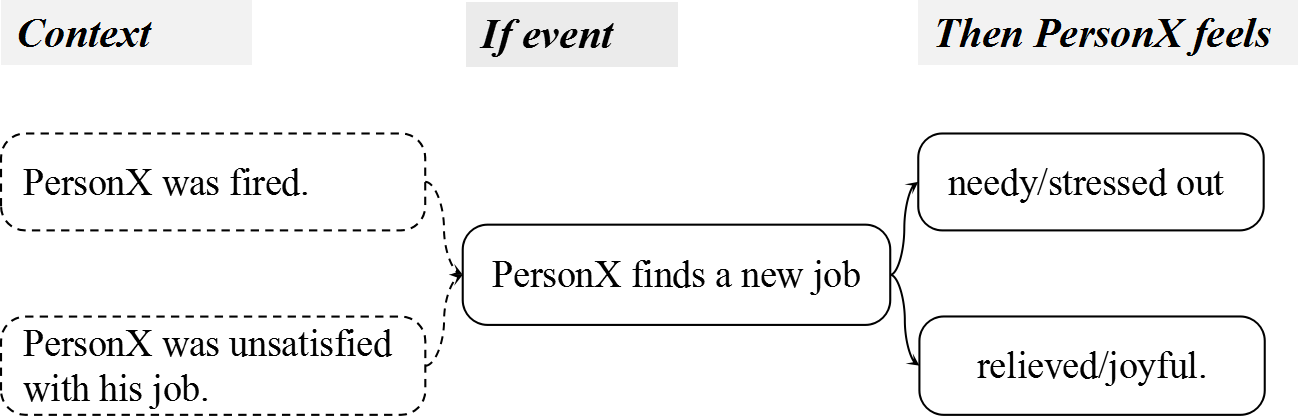}
  \caption{A illustration of two challenging problems in If-Then reasoning. (a) Given an observed event, the feelings about this event could be multiple. (b) Background knowledge is need for generating reasonable inferences, which is absent in the dataset (marked by dashed lines).}
  \label{figure:1}
\end{figure}

However, there still remains two challenging problems. First, as illustrated in Figure~\ref{figure:1}, given an event ``PersonX finds a new job'', the plausible feeling of PersonX about that event could be multiple (such as \emph{``needy/stressed out''} and \emph{``relieved/joyful''}). Previous work showed that for the one-to-many problem, conventional RNN-based encoder-decoder models tend to generate generic responses, rather than meaningful and specific answers \cite{li2016diversity,serban2016building}.

Second, as a commonsense reasoning problem, rich background knowledge is necessary for generating reasonable inferences. For example, as shown in Figure~\ref{figure:1}, the feeling of PersonX upon the event ``PersonX finds a new job'' could be multiple. However, after given a context ``\emph{PersonX was fired}'', the plausible inferences would be narrowed down to ``\emph{needy}'' or ``\emph{stressed out}''.

To better solve these problems, we propose a context-aware variational autoencoder (CWVAE) together with a two-stage training procedure. Variational Autoencoder (VAE) based models have shown great potential in modeling the one-to-many problem and generate diversified inferences \cite{bowman2015generating,zhao2017learning}. 

In addition to the traditional VAE structure, we introduces an extra context-aware latent variable in CWVAE to learn the event background knowledge. In the pretrain stage, CWVAE is trained on an auxiliary dataset (consists of three narrative story corpora and contains rich event background knowledge), to learn the event background information by using the context-aware latent variable. Subsequently, in the finetune stage, CWVAE is trained on the task-specific dataset to adapt the event background information to each specific aspect of If-Then inferential target (e.g., intents, reactions, etc.).

Experiments on the Event2Mind and Atomic dataset show that our proposed approach outperforms baseline methods in both the accuracy and diversity of inferences. The code is released at \url{https://github.com/sjcfr/CWVAE}.


\section{Background}


Before specifically describing two dataset ---- Event2Mind and Atomic used in this paper as well as the If-Then reasoning task, for clarity, we define the following terminologies:

\textbf{Base event:} the prerequisite event in If-Then reasoning, organized as a verb phrase with a predicate and its arguments, such as the event ``PersonX finds a new job'' shown in Figure~\ref{figure:1}.

\textbf{Inference dimension:} a particular If-Then reasoning type, e.g., intents, effects of the base event. Details are shown in Table~\ref{table:e_struc} and Table~\ref{table:a_struc}.

\textbf{Target:} the inferential results. For example, as shown in Figure~\ref{figure:1}, given a base event ``PersonX finds a new job'' and one inference dimension ``xReact'', the targets could be ``\emph{relieved}'' or ``\emph{needy}''. Notice that each inference dimension can have multiple targets.

\noindent \textbf{Event2Mind Dataset} contains 25K base events and 300K targets, annotated through crowdsourcing. Event2Mind is organized in a hierarchical form: each base event has three types of inference dimensions, and given a base event, under one of inference dimensions, several targets may simultaneously exist. Table~\ref{table:e_struc} shows the (base event-inference dimension-target) hierarchical structure through an example from Event2Mind.


\noindent \textbf{Atomic Dataset} Inspired by Event2Mind, the Atomic dataset shares the same hierarchical structure as Event2Mind, while scales up the size of dataset and expands the scope to nine types of inference dimensions. 
Table~\ref{table:a_struc} shows the (base event-inference dimension-target) hierarchical structure through an example from Atomic. Though Atomic covers the inference dimensions of Event2Mind, the base event collection of Event2Mind is nonidentical to that of Atomic.  

\begin{table}[]
\footnotesize
\centering
\begin{tabular}{L{1.7cm}L{1.1cm}L{3.7cm}}

\hline
Base event & Inference Dim. & Target \\
\hline
\multirow{3}*{\tabincell{l}{\\ PersonX \\writes PersonY \\ a letter}} & xIntent & \tabincell{l}{to send a message, \\ express themself} \\ \cline{2-3}
~ & xReact & \tabincell{l}{nervous,\\ thoughtful}  \\ \cline{2-3}
~ & oReact &  \tabincell{l}{indifferent,\\ receptive} \\ 
\hline
\end{tabular}
\caption{Hierarchical structure of Event2Mind dataset. For specific inference dimensions, ``x'' and ``o'' refers to PersonX and others respectively.}
\label{table:e_struc}
\end{table}

\begin{table}[]
\footnotesize
\centering
\begin{tabular}{L{1.7cm}L{1.1cm}L{3.7cm}}

\hline
Base event & Inference Dim. & Target \\
\hline
\multirow{9}*{\tabincell{l}{\\ \\ \\ \\ \\ \\PersonX \\adopts a child}} & xIntent & \tabincell{l}{to help another person,\\ to have a child} \\ \cline{2-3}
~ & xNeed  & \tabincell{l}{to visit adoption agency,\\ to be approved for adoption} \\ \cline{2-3}
~ & xAttr   & \tabincell{l}{compassionate,\\ generous} \\ \cline{2-3}
~ & xEffect  & \tabincell{l}{becomes a parent,\\ gains love and companionship} \\ \cline{2-3}
~ & xWant  & \tabincell{l}{take child home,\\ buy child clothes} \\ \cline{2-3}
~ & xReact & \tabincell{l}{happy,\\ caring}  \\ \cline{2-3}
~ & oReact &  \tabincell{l}{has a parent,\\ receives love and affection} \\ \cline{2-3}
~ & oWant  & \tabincell{l}{try on new clothes,\\ to have a family}   \\ \cline{2-3}
~ & oEffect & \tabincell{l}{has a parent,\\ Receives love and affection}   \\
\hline
\end{tabular}
\caption{Hierarchical structure of Atomic dataset. For specific inference dimensions, ``x'' and ``o'' refers to PersonX and others respectively.}
\label{table:a_struc}
\end{table}

\noindent \textbf{Problem Definition} The If-Then reasoning task could be formally defined as a conditional one-to-many generation problem: given a base event $x$ and one inference dimension $d$, the model is required to generate targets $y=f(x, d)$ as close to the ground truths as possible. Both $x$ and $y$ consist of sequence of words: $x=\{x_1,\dots, x_{m}\}$, and $y=\{y_1,\dots, y_{n}\}$, where $m$ and $n$ denotes the length of $x$ and $y$, respectively.

\begin{figure}
  \centering
  \includegraphics[width=0.55\linewidth]{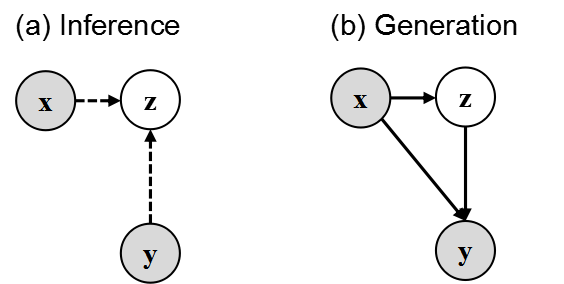}
  \caption{Illustration of inference and generation process of CVAE in a directed graph. Dashed lines represent the inference of $z$. Solid lines represent the generation process.}
  \label{figure:cvae}
\end{figure}

\noindent \textbf{Conditional Variational Autoencoder} The variational autoencoder (VAE) defines a generative framework suited for one-to-many generation problem \cite{kingma2014auto}. While conditional variational autoencoder (CVAE) \cite{sohn2015learning} is an extension of VAE on the conditional generation problem. As shown in Figure~\ref{figure:cvae}~(a), CVAE characterizes the conditional one-to-many generation problem using three random variables: event $x$, target $y$ and a latent variable $z$, which is used for modeling the latent distribution of semantic over targets given an event.
Hence, under a certain inference dimension, with regard to the latent semantic variable $z$, the conditional generation problem could be expressed as $p(y|x)=\int p(y|x,z)p(z|x)dz$. CVAE models $p(y|x,z)$ and $p(z|x)$ using deep neural networks (parameterized by $\theta$) $p_{\theta}(y|x,z)$ and $p_{\theta}(z|x)$. Then as illustrated in Figure~\ref{figure:cvae}~(b), $y$ could be generated from $x$ and $z$.

CVAE is trained to maximize the conditional likelihood $p(y|x)$, which involves an intractable marginalization over the latent variable $z$. Instead, following \citeauthor{kingma2014auto} (\citeyear{kingma2014auto}), a practical way is to introduce another deep network (parameterized by $\phi$) $q_{\phi}(z|x,y)$ to approximate the true posterior distribution $p(z|x,y)$ and maximize the evidence lower bound (ELBO) of the log-likelihood function:
\begin{align}
\vspace*{-1\baselineskip}
  \begin{split}
     L^{ELBO}(\theta, \phi)=&\mathbb{E}_{q_{\phi}(z|x, y)}\mathrm{log}(p_{\theta}(y|x, z))-\\
    &\mathrm{KL}(q_{\phi}(z|x, y)||p_{\theta}(z|x))\\
    &\leq \mathrm{log}p(y|x)
  \end{split}
\label{eq:2}
\end{align}

Therefore, CVAE is composed of three neural networks in general. We refer to $p_{\theta}(z|x)$ as a prior network, $q_{\phi}(z|x,y)$ as a recognition network, and $p_{\theta}(y|x,z)$ as a neural decoder. 

\begin{figure}
  \centering
  \includegraphics[width=0.95\linewidth]{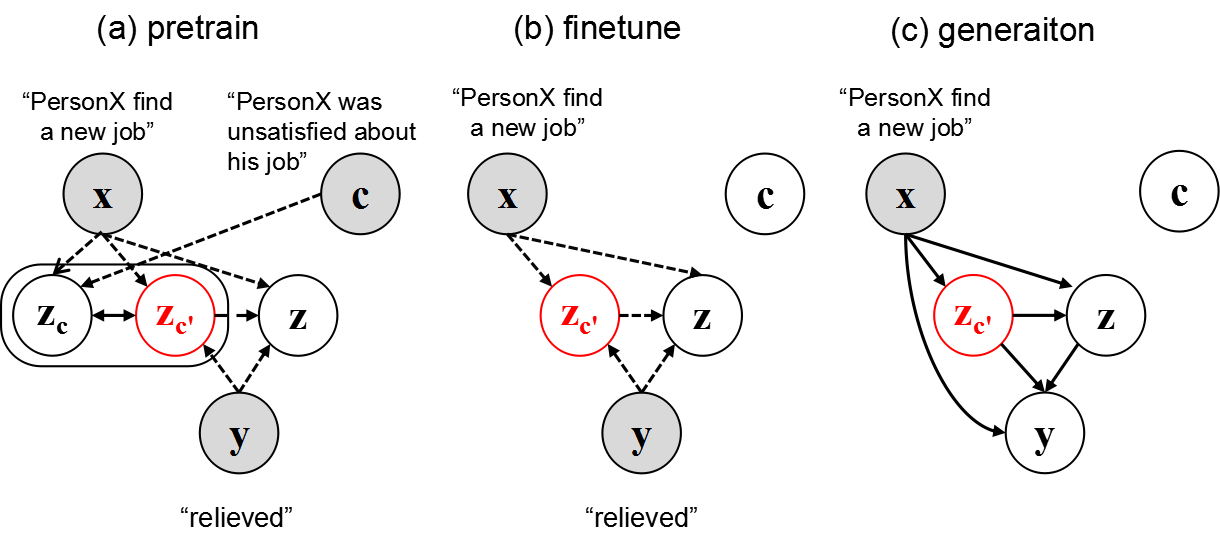}
  \caption{Illustration of pretrain, finetune and generation process of CWVAE in a directed graph. Dashed lines represent the inference of $z$, $z_c$ and $z_{c'}$. Solid lines represent the generation process. Red circle denotes the context-aware latent variable.}
  \label{figure:3}
\end{figure}


\section{Context-aware Variational Autoencoder}

Traditional CVAE can model the {\small\verb|event-target|} relation. In other words, given an observed event, CVAE can generate its corresponding targets. While in this paper we model the If-Then reasoning as a {\small\verb|[(background), event]-target|} process. It means that in addition to the observed event, we also want to involve the event background knowledge (which can be learned from event contexts) to generate the reasonable targets. 

To this end, we propose a context-aware variational autoencoder (CWVAE), with two additional latent variables: a context-acquiring latent variable $z_c$ to directly acquire context information, and a context-aware latent variable $z_{c'}$ to learn background knowledge from $z_c$, as shown in Figure~\ref{figure:3}~(a). However, the event context information is absent in the Event2Mind and  Atomic dataset. To learn from the external event context information, we design the following two-stage training procedure for CWVAE.



\noindent \textbf{Pretrain: Learning Event Background Knowledge from Auxiliary Dataset}
In the pretrain stage, CWVAE is trained on three narrative story corpora with rich event context information. As shown in Figure~\ref{figure:3}~(a), context-acquiring latent variable $z_c$ is directly conditioned on the context $c$. Hence, $z_c$ could be employed for acquiring background knowledge from event contexts. Then, we minimize the distance between $z_c$ and the context-aware latent variable $z_{c'}$, by which the event background knowledge is transferred from $z_c$ to $z_{c'}$.

\noindent \textbf{Finetune: Adapt Event Background Knowledge to Each Inference Dimension}
In the finetune stage, as shown in Figure~\ref{figure:3}~(b), CWVAE is trained on the Event2Mind and Atomic dataset without the event context information. Pretrained CWVAE is finetuned to learn the specific inferential knowledge of each inference dimension.
After the training procedure, as shown in Figure~\ref{figure:3}~(c), samples of $z$ is generated based on $x$ and samples of $z_{c'}$, where $z_{c'}$ contains rich event background knowledge helpful for If-Then reasoning. 

\subsection{Architecture of CWVAE}
\begin{figure}
  \centering
  \includegraphics[width=1.05\linewidth]{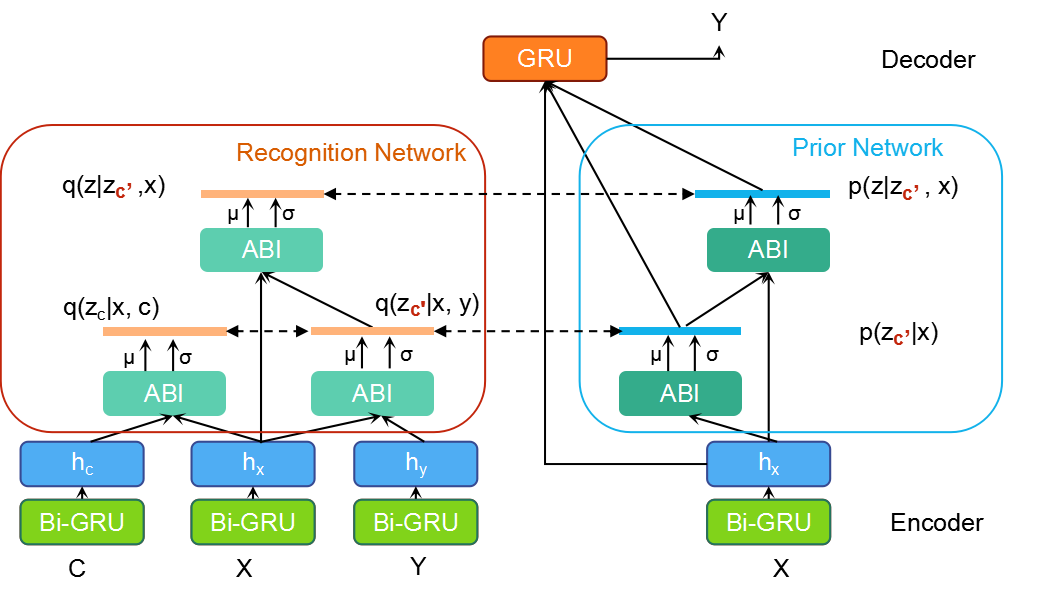}
  \caption{Architecture of CWVAE. We mark Neural encoder in green, prior network in blue, recognition network in brown and neural decoder in orange, respectively.}
  \label{figure:4}
\end{figure}

As shown in Figure~\ref{figure:4}, CWVAE is mainly composed of four parts: 
a neural encoder that provides distributed representations of base events/targets, a recognition network for inferring $q_{\phi}(z_{c'}|x,y)$, $q_{\phi}(z_c|x,c)$ and $q_{\phi}(z|z_{c'}, x)$, a prior network for modeling $p_{\theta}(z_{c'}|x)$ and $p_{\theta}(z|x, z_{c'})$, and a neural decoder that integrates the information from $z$ and $z_{c'}$ to generate targets.

\noindent \textbf{Neural Encoder} We employ a bidirectional GRU as neural encoder, which encodes context $c$, event $x$ and target $y$ into distributed representations $h^c=\{h_1^c,\dots,h_{l_c}^c\}$, $h^x=\{h_1^x,\dots,h_{l_x}^x\}$ and $h^y=\{h_1^y,\dots,h_{l_y}^y\}$, where $l_c$, $l_x$ and $l_y$ is the length of $c$, $x$ and $y$, respectively.

\noindent \textbf{Recognition Network}
The recognition network models $q_{\phi}(z|x,y)$, $q_{\phi}(z_c|x,c)$, $q_{\phi}(z|z_{c'}, x)$ based on $h^x$, $h^y$ and $h^c$.

Following traditional VAE, the above-mentioned three distributions are assumed to be multivariate Gaussian distribution with a diagonal covariance structure:
\begin{equation}
\begin{split}
  &q_{\phi}(z_{c}|x,c) \sim N(\mu_{z_{c}}(x,c), \sigma_{z_{c}}(x,c)I) \\
  &q_{\phi}(z_{c'}|x,y) \sim N(\mu_{z_{c'}}(x,y), \sigma_{z_{c'}}(x,y)I) \\
  &q_{\phi}(z|x,y) \sim N(\mu_{z}(x,y), \sigma_z(x,y)I)
\end{split}
\end{equation}
where $\mu$ denotes the mean of the distribution, $\sigma$ denotes the standard deviation of the distribution, and $I$ denotes the identity matrix.

Given $h^x$, $h^y$ and $h^c$, we propose a novel attention-based inferer (ABI) module to estimate the mean and standard deviation of $q_{\phi}(z_{c}|x,c)$, $q_{\phi}(z_{c'}|x,y)$ and $q_{\phi}(z|x,y)$:
\begin{equation}
\begin{split}
&\mu_{z_c}, \sigma_z=\mathrm{ABI}_c(h^c, h^x) \\
&\mu_{z_{c'}}, \sigma_z=\mathrm{ABI}_{c'}(h^y, h^x) \\
&\mu_{z}, \sigma_z=\mathrm{ABI}_z(z_{c'}, h^x)
\end{split}
\end{equation}

Briefly, through the attention mechanism, ABI can capture the semantic interaction between input sequences, and estimate the parameters of distributions based on it. We will introduce the specific structure of ABI in below.

\noindent \textbf{Prior Network}
Prior Network models $p_{\theta}(z_{c'}|x)$ and $p_{\theta}(z|x, z_{c'})$ based on $h^x$. The distribution of $p_{\theta}(z_{c'}|x)$ and $p_{\theta}(z|x, z_{c'})$ are still assumed to be multivariate Gaussian, whereas the parameters are different:
\begin{equation}
\begin{split}
&p_{\theta}(z|x) \sim N(\mu_{z}^{'}(x), \sigma_z^{'}(x)I) \\
&p_{\theta}(z_{c'}|x) \sim N(\mu^{'}_{z_{c'}}(x), \sigma^{'}_{z_{c'}}(x)I)
\end{split}
\end{equation}
where $\mu^{'}$ denotes the mean of the distribution, $\sigma^{'}$ denotes the standard deviation of the distribution and $I$ denotes the identity matrix.

Then the attention-based inferer module is still employed to estimate parameters of distributions:
\begin{equation}
\begin{split}
&\mu^{'}_{z_{c'}}, \sigma^{'}_z=\mathrm{ABI}_{c'}^{'}(h^x, h^x) \\
&\mu^{'}_{z}, \sigma^{'}_z=\mathrm{ABI}_z^{'}(z_{c'}, h^x)
\end{split}
\end{equation}

\noindent \textbf{Neural Decoder} Given the base event $x$, the semantic latent variable $z$, and the context-aware latent variable $z_{c'}$, the neural decoder defines the generation probability of $y$ as following:
\begin{equation}
p(y|x,z,z_{c'})=\prod_{j=1}^{n}p(y_j|y<j, z, z_{c'}, x)
\end{equation}
where $p(y_j|y<j, z, z_{c'}, x)=g(y_{j-1}, s_{j-1}, e_j)$, $g(\cdot)$ is an attention-based feed forward model, $e_j=\sum_i \alpha_{ji}h_i^{x}$ is the context vector and $s_{j-1}$ is the hidden state of the decoder. We obtain $g(\cdot)$ and $e_j$ the same way as \citeauthor{bahdanau2014neural} (\citeyear{bahdanau2014neural}). Whereas our decoder differs from \citeauthor{bahdanau2014neural} (\citeyear{bahdanau2014neural}) in that our model integrates the context-aware latent variable $z_{c'}$ and semantic latent variable $z$ in the computation of $s_j=\mathrm{GRU}([E_{yj};s_{j-1},z,z_{c'}])$, where $E_{yj}$ is the word embeddings of target words.

Note that through concatenating $z$ and $z_{c'}$ with $E_{yj}$ and $s_{j-1}$, $s_j$ could be affected by context-aware latent variable $z_{c'}$ and semantic latent variable $z$. This allows model to directly access to the event background knowledge from $z_{c'}$. In addition, the randomness of $z$ and $z_{c'}$ would increase the diversity of model generation.

\noindent \textbf{Attention-based Inferer} Attention mechanism has shown strong ability in capturing semantic interactions \cite{gong2017natural}. Inspired by the co-attention mechanism \cite{parikh2016decomposable}, we propose an attention-based inferer (ABI) to estimate 
the mean and standard deviation of a distribution belongs to $p_{\theta}(\cdot)$ or $q_{\phi}(\cdot)$ by capturing semantic interactions of input sequences.

Specifically, given two input sequences (e.g., representations of contexts and events) $a=\{a_1,\dots,a_{l_a}\}$ and $b=\{b_1,\dots,b_{l_b}\}$ with length $l_a$ and $l_b$, we first obtain the attention scores from each side through:
\begin{equation}
  \small
  \begin{split}
  \gamma_t^a=\frac{\mathrm{exp}[(W_a h_t^a)^T {(W_b h_i^b)}]}{\sum_i \mathrm{exp}[(W_a h_t^a)^T {(W_b h_i^b)}]} \\
  \gamma_t^b=\frac{\mathrm{exp}[(W_a h_i^a)^T {(W_b h_t^b)}]}{\sum_i \mathrm{exp}[(W_a h_i^a)^T {(W_b h_t^b)}]}
  \end{split}
\end{equation}
where $W_a \in \mathbb{R}^{d\times d_a}$ and $W_b \in \mathbb{R}^{d\times d_b}$ are parameter weights.

With these attention scores, the context vectors of both sequences are given by:
\begin{equation}
\begin{split}
  &c_t^a=a\gamma_t^a \\
  &c_t^b=b\gamma_t^b \\
\end{split}
\end{equation}

Then we perform a mean pooling operation on context vectors of both sequences:
\begin{equation}
\small
\begin{split}
  &\bar{c^a}=\frac{1}{l_a}\sum_{t=1}^{l_a}c_t^a \\
  &\bar{c^b}=\frac{1}{l_b}\sum_{t=1}^{l_b}c_t^b \\
\end{split}
\end{equation}

To obtain the mean and standard deviation, the pooled context vectors $\bar{c^a}$ and $\bar{c^b}$ which carry semantic interaction between two sequences, are concatenated and projected into a latent semantic space through a nonlinear transformation:
\begin{equation}
  h_z=\mathrm{tanh}(W[\bar{c}^a; \bar{c}^b] + b_z)
\end{equation}

Finally the mean and standard deviation are generated through a nonlinear transformation over $h_z$:
\begin{equation}
\begin{split}
  &\mu=W_{\mu}h_z+b_{\mu} \\
  &\sigma=\mathrm{softplus}(W_{\sigma}h_z+b_{\sigma}) \\
\end{split}
\end{equation}

\subsection{Optimizing}

With the incorporation of $z_{c'}$, the original loglikelihood could be decomposed as:
\begin{equation}
  \mathrm{log}p(y|x)=\iint \ p(y|z, z_{c'}, x) dzdz_{c'}
\end{equation}

Then following traditional CVAE, the ELBO of CWVAE is defined as follows:
\begin{equation}
\small
\begin{split}
&L^{ELBO}(\theta,\phi)=\\
&\qquad \underbrace{\mathbb{E}_{q_{\phi}(z|x, z_{c'})q_{\phi}(z_{c'}|x,y)}\mathrm{log}p_{\theta}(y|x, z, z_{c'})}_{\text{Reconstruction Loss}}\\
&-\underbrace{\int q_{\phi}(z|x,y) \mathrm{KL}(q_{\phi}(z_{c'}|x,y)||p_{\theta}(z_{c'}|x)) d{z}}_{\text{KL term}} \\
&-\underbrace{\int q_{\phi}(z_{c'}|x,y) \mathrm{KL}(q_{\phi}(z|x, z_{c'})||p_{\theta}(z|x,z_{c'})) d{z_{c'}}}_{\text{KL term}}
\end{split}
\end{equation}
which is the objective function at the finetune stage.

While in the pretrain stage, as we aim to learn background knowledge through minimizing the distance between $z_c$ and $z_{c'}$, in addition to $L^{ELBO}$, a context-aware regulation term is introduced:
\begin{equation}
\small
\begin{split}
L^{PT}(\theta,\phi)=&L^{ELBO}(\theta,\phi) \hfill \\
 -&\underbrace{\lambda \int \mathrm{KL}(q_{\phi}(z_c|x,c)||q_{\phi}(z_{c'}|x,y)) d{z_{c'}}}_{\text{context-aware regulation}}
\end{split}
\end{equation}
where the context aware regularization term is the KL distance between $z$ and $z_{c'}$. Through minimizing the context aware regularization term, we aim to pass event context knowledge from $z_c$ to the context aware latent variable $z_{c'}$.

\subsection{Training Details}

To test the performance of CWVAE, we split the Event2Mind and Atomic dataset into training, development and test sets (80\%, 10\%, 10\%) in the same way as \citeauthor{rashkin2018event2mind} (\citeyear{rashkin2018event2mind}) and \citeauthor{sap2018atomic}  (\citeyear{sap2018atomic}), respectively. 

We initialize the embedding layer from 300d GloVe word embeddings. The neural encoder is chosen to be biGRU with 300 hidden units. For the ABI module, size of $W_a$ and $W_b$ is set to be $100 \times d_a$ and $100 \times d_b$ respectively. The dimension of $z_c$, $z_{c'}$ and $z$ is all set as 40. The neural decoder is set to be GRU with 300d hidden state. Regulation coefficient $\lambda$ of context-aware regulation term is set to be 0.1. Models are trained using an Adam optimizer \cite{kinga2015method} with a learning rate of 0.001.

\section{Experiments}
\subsection{Auxiliary Dataset}

\begin{table*}[]
\footnotesize
\centering
\begin{tabular}{c|c|c}
\hline
Context & Event & Inference Target \\
\hline
 \tabincell{l}{\textcircled{1} jason had been really stressed out at work.\\ \textcircled{2} he decided he needed a different kind of job. \\ \textcircled{3} jason applied for a job in a different field.} &	\textcircled{4} he got the job . & \textcircled{5} jason was much happier at his new job . \\
\hline
\end{tabular}
\caption{An example for the construction of auxiliary dataset. For a five-sentence-paragraph, the first three sentences are taken as event context, while the fourth and fifth sentence is taken as base event and target respectively.}
\label{table:construct_ad}
\end{table*}

The auxiliary dataset is built upon three human-written story corpora: ROCStories \cite{mostafazadeh2016corpus}, VIST \cite{huang2016visual} and WritingPrompts \cite{fan2018hierarchical}. ROCStories and VIST are composed of short stories with five sentences. We filter out stories of more than 1,000 words in WritingPrompts, and cut the remaining stories into five-sentence-paragraphs.

For each five-sentence-paragraph, we define the first three sentences as contexts of the base event, the fourth sentence as the base event, and the fifth sentence as the inference target. For example, as shown in Table~\ref{table:construct_ad}, the first three sentences describe a context that Jason was unsatisfied about his job and applied for a new job. Hence, after happening the event ``he got the job'', a plausible react about the event could be ``jason was much happier at his new job''. In total, the auxiliary dataset contains 192,316 $(context, event, target)$ triples.

\subsection{Baselines}

We compared our proposed model with the following four baseline methods:

\begin{list}{\labelitemi}{\leftmargin=1em}
    \setlength{\topmargin}{0pt}
    \setlength{\itemsep}{0em}
    \setlength{\parskip}{0pt}
    \setlength{\parsep}{0pt}
  \item \textbf{RNN-based Seq2Seq} proposed by \citeauthor{sap2018atomic} (\citeyear{sap2018atomic}) for the If-Then reasoning on Atomic.
  \item \textbf{Variational Seq2Seq} combines a latent variable with the encoder-decoder structure through converting the last hidden state of RNN encoder into a Gaussian distributed latent variable \cite{bowman2015generating}.
  \item \textbf{VRNMT} 
  Propose by \citeauthor{su2018variational} (\citeyear{su2018variational}), VRNMT combines CVAE with attention-based encoder-decoder framework through introduces a latent variable to model the semantic distribution of targets. 
  \item \textbf{CWVAE-Unpretrained} refers to the CWVAE model without the pretrain stage.

\end{list}
Note that, for each baseline method, we train distinct models for each distinct inference dimension, respectively.

\begin{table}
    \centering
    \footnotesize
  \setlength{\abovecaptionskip}{0.1cm}
\begin{tabular}{L{0.7cm}|L{2.9cm}|C{0.65cm}C{0.65cm}C{0.65cm}}
  \hline
  Metric & Methods & xIntent & xReact & oReact \\
  \hline
\multirow{5}*{PPL} & RNN-based Seq2Seq & 44.12  & 29.18  & 14.08   \\
  ~ & Variational Seq2Seq & 42.06 & 28.22 & 12.62  \\
  ~ & VRNMT & 33.45 &	25.54 &	11.93   \\ \cline{2-5}
  ~ & CWVAE-Unpretrained & 31.32 & 24.07 & 11.37  \\
  ~ & CWVAE & \textbf{29.23} & \textbf{23.17} & \textbf{11.04}  \\
  \hline
  \hline
  \multirow{5}*{BLEU} & RNN-based Seq2Seq & 2.75  & 2.11  & 5.18   \\
  ~ & Variational Seq2Seq & 2.84 & 2.43 & 2.08  \\
  ~ & VRNMT & 4.81 & 3.94 & 6.61  \\ \cline{2-5}
  ~ & CWVAE-Unpretrained & 7.36 & 5.52 &  5.33 \\
  ~ & CWVAE & \textbf{12.98} & \textbf{5.65} & \textbf{6.97}  \\
  \hline
  \end{tabular}
    \caption{Average perplexity and BLEU score (reported in percentages) for the top 10 generations under each inference dimension of Event2Mind. The the best result for each dimension is emboldened.}
    \label{table:accu_e}
\end{table}{}

\begin{table}[]
    \centering
    \footnotesize
    \setlength{\abovecaptionskip}{0.1cm}
  \setlength{\belowcaptionskip}{-0.3cm}
\begin{tabular}{L{0.7cm}|L{2.9cm}|C{0.65cm}C{0.65cm}C{0.65cm}}
  \hline
  Metric & Methods & xIntent & xReact & oReact \\
  \hline
  \multirow{5}*{dist-1} & RNN-based Seq2Seq & 0.0002 & 0.0002 & 0.0001   \\
  ~ & Variational Seq2Seq & 0.0006 & 0.0003 & 0.0001 \\
  ~ & VRNMT & 0.0002 & 0.0002 &	0.0003  \\ \cline{2-5}
  ~ & CWVAE-Unpretrained & 0.0023 & 0.0017 & 0.0004  \\
  ~ & CWVAE & \textbf{0.0052} & \textbf{0.0033} & \textbf{0.0025}  \\
  \hline
  \hline
  \multirow{5}*{dist-2} & RNN-based Seq2Seq & 0.0005  & 0.0002  & 0.0002   \\
  ~ & Variational Seq2Seq & 0.0014 & 0.0002 & 0.0001  \\
  ~ & VRNMT & 0.0005 &	0.0003 &	0.0001  \\ \cline{2-5}
  ~ & CWVAE-Unpretrained & 0.0061 & 0.0040 & 0.0013  \\
  ~ & CWVAE & \textbf{0.0146} & \textbf{0.0099} & \textbf{0.0063}  \\
  \hline
  \end{tabular}
    \caption{Distinct-1 and distinct-2 scores for the top 10 generations under each inference dimension of Event2Mind. The the best result for each dimension is emboldened.}
    \label{table:div_e}
\end{table}{}

\begin{table*}[]
    \centering
    \footnotesize
  \setlength{\abovecaptionskip}{0.1cm}
\begin{tabular}{C{0.85cm}|C{3.15cm}|C{0.81cm}C{0.81cm}C{0.81cm}C{0.81cm}C{0.81cm}C{0.81cm}C{0.81cm}C{0.81cm}C{0.81cm}}
  \hline
  Metric & Methods & xIntent & xNeed & xAttr & xEffect & xReact & xWant & oWant & oReact & oEffect \\
  \hline
  \multirow{5}*{PPL} & RNN-based Seq2Seq & 22.54  & 24.69  & 33.54  & 65.13 & 29.52 & 26.63 & 16.76 & 14.99 & 35.17 \\
  ~ & Variational Seq2Seq & 26.48 & 28.31 & 33.00 &	68.62 & 29.93 & 29.50 & 16.98 & 14.25 & 34.20 \\
  ~ & VRNMT & 21.04 & 24.28 &	24.87 &	61.05 &	26.62 &	28.57 &	14.45 &	14.86 &	30.12 \\ \cline{2-11}
  ~ & CWVAE-Unpretrained & 20.73 & 23.72 & 25.80 & 60.62 & 25.75 &	26.71 &	15.93 & 12.82 &	32.00 \\
  ~ & CWVAE & \textbf{15.93} & \textbf{20.32} & \textbf{23.85} & \textbf{50.74} & \textbf{21.39} & \textbf{24.02} & \textbf{14.02} &	\textbf{11.70} & \textbf{29.13} \\
  \hline
  \hline
  \multirow{5}*{BLEU} & RNN-based Seq2Seq & 8.17  & 12.35  & 2.96  & 5.26 & 3.43 & 13.44 & 7.08 & 4.09 & 6.42 \\
  ~ & Variational Seq2Seq & 8.31 & 12.05 & 2.13 &	6.07 & 2.52 & 11.71 & 7.40 & 4.08 & 6.38 \\
  ~ & VRNMT & 9.52 &	13.35 &	4.87 &	4.42 &	7.64 &	9.80 &	10.79 &	5.28 &	13.71 \\ \cline{2-11}
  ~ & CWVAE-Unpretrained & 11.37 & 14.64 & 4.07 & 14.11 & 7.86	& 12.70 & 12.09 & 8.16 &	\textbf{14.93} \\
  ~ & CWVAE & \textbf{12.12} & \textbf{15.67} & \textbf{5.63} & \textbf{14.64} & \textbf{8.13} & \textbf{15.01} & \textbf{13.83}  &	\textbf{8.58} & 11.63 \\
  \hline
  \end{tabular}
    \caption{Average perplexity and BLEU scores (reported in percentages) for the top 10 generations under each inference dimension of Atomic. The the best result for each dimension is emboldened.}
    \label{table:accu_a}
\end{table*}{}

\begin{table*}[]
    \centering
    \footnotesize
  \setlength{\abovecaptionskip}{0.1cm}
  \setlength{\belowcaptionskip}{-0.3cm}
\begin{tabular}{C{0.85cm}|C{3.15cm}|C{0.81cm}C{0.81cm}C{0.81cm}C{0.81cm}C{0.81cm}C{0.81cm}C{0.81cm}C{0.81cm}C{0.81cm}}
  \hline
  Metric & Methods & xIntent & xNeed & xAttr & xEffect & xReact & xWant & oWant & oReact & oEffect \\
  \hline
  \multirow{5}*{dist-1} & RNN-based Seq2Seq & 0.0012  & 0.0029  & 0.0004  & 0.0019 & 0.0001 & 0.0022 & 0.0006 & 0.0001 & 0.0006 \\
  ~ & Variational Seq2Seq & 0.0006 & 0.0018 & 0.0002 &	0.0002 & 0.0001 & 0.0013 & 0.0007 & 0.0001 & 0.0002 \\
  ~ & VRNMT & 0.0002 &	0.0001 & 0.0053 &	0.0005 &	0.0018 &	0.0022 &	0.0005 &	0.0001 & 0.0004 \\ \cline{2-11}
  ~ & CWVAE-Unpretrained & 0.0019 & 0.0036 & 0.0119 & \textbf{0.0046} & 0.0021 &	0.0013 &	0.0018 & 0.0005 &	0.0006 \\
  ~ & CWVAE & \textbf{0.0055} & \textbf{0.0045} & \textbf{0.0142} & 0.0028 & \textbf{0.0043} & \textbf{0.0040} & \textbf{0.0021} &	\textbf{0.0030} & \textbf{0.0033} \\
  \hline
  \hline
  \multirow{5}*{dist-2} & RNN-based Seq2Seq & 0.0036 & 0.0081  & 0.0002  & 0.0018 & 0.0002 & 0.0006 & 0.0013 & 0.0001 & 0.0011 \\
  ~ & Variational Seq2Seq & 0.0013 & 0.0042 & 0.0001 &	0.0003 & 0.0002 & 0.0026 & 0.0002 & 0.0003 & 0.0006 \\
  ~ & VRNMT & 0.0002 & 0.0011 &	0.0002 & 0.0005 & 0.0001 & 0.0034 &	0.0005 &	0.0001 &	0.0004 \\ \cline{2-11}
  ~ & CWVAE-Unpretrained & 0.0060 & 0.0088 & 0.0136 & \textbf{0.0113} & 0.0043 &	0.0029 &	0.0041 & 0.0011 &	0.0009 \\
  ~ & CWVAE & \textbf{0.0162} & \textbf{0.0112} & \textbf{0.0146} & 0.0072 & \textbf{0.0013} & \textbf{0.0107} & \textbf{0.0044} &	\textbf{0.0068} & \textbf{0.0093} \\
  \hline
\end{tabular}
    \caption{Distinct-1 and distinct-2 scores for the top 10 generations under each inference dimension of Atomic. The the best result for each dimension is emboldened.}
    \label{table:div_a}
\end{table*}{}

\subsection{Evaluation Metrics}
\paragraph{Automatic Evaluation} We first compare the perplexity of CWVAE with baseline methods. Perplexity measures the probability of model to regenerate the exact targets, which is particular suitable for evaluating the model performance on one-to-many problem \cite{serban2017hierarchical}. Further, we employ BLEU score to evaluate the accuracy of generations \cite{papineni2002bleu}, and the number of distinct n-gram to evaluate the diversity of generations \cite{li2016diversity}. The distinct is normalized to $[0, 1]$ by dividing the total number of generated tokens.

\paragraph{Human Evaluation} Since automatic evaluation of generations is still a challenging task \cite{liu2016not}, we also conduct human evaluations on the model performance. Five human experts are employed to evaluate the \emph{coherence}, \emph{diversity} and \emph{fluency} of generated targets. Experts are asked to vote for if a generation is fluent or coherent for each generated target, and give a 1-5 score for the diversity of generations. For both Event2Mind and Atomic datasets, 100 events are randomly selected from the test set. For each method, top 10 generated targets of each base event are used for evaluation. Finally we report three overall averaged scores of \emph{coherence}, \emph{diversity} and \emph{fluency} on both datasets, respectively.


\subsection{Overall Results}

We list the perplexity and BLEU score of CWVAE and baseline methods on Event2Mind and Atomic in Table~\ref{table:accu_e} and Table~\ref{table:accu_a}, respectively, and show the distinct-1 and distinct-2 score on Event2Mind and Atomic in Table~\ref{table:div_e} and Table~\ref{table:div_a}, respectively. We find that:

(1) As shown in Table~\ref{table:div_e} and Table~\ref{table:div_a}, comparison between RNN-based Seq2Seq and variational-based methods, including Variational Seq2Seq, VRNMT, CWVAE-unpretrained and CWVAE shows that, variational-based methods could increase the diversity of generations. This confirms one of our motivations that variational-based methods could capture the latent semantic distribution within targets and increase the diversity of If-Then reasoning.


(2) Comparing CWVAE-unpretrained with other baseline methods shows that, in general CWVAE improves the accuracy and diversity on both dataset. These results indicate the efficiency of CWVAE in capturing the latent semantic distribution of targets, and generate more reasonable inferential results.

(3) Comparison between CWVAE and CWVAE-unpretrained shows that the pretrain stage could enhance the performance of CWVAE in both the accuracy and diversity. This is mainly because event knowledge could offer the guidance for If-Then reasoning. In the pretrain stage, CWVAE could capture the event background knowledge through context-aware latent variable, and such knowledge could be be adapted to our task through the fintune stage.



\begin{table}[]
  \setlength{\abovecaptionskip}{0.1cm}
\footnotesize
\centering
\begin{tabular}{L{3cm}|L{1cm}C{1cm}C{1cm}}
  \hline
  Methods & Coherence & Diversity & Fluency  \\
  \hline
  RNN-based Seq2Seq & 0.28 & 2.03 & 0.73 \\
  Variational Seq2Seq & 0.33 & 1.67 & 0.92 \\
  VRNMT & 0.32 & 2.60 & 0.83 \\
  \hline
  \hline
  CWVAE-Unpretrained & 0.36 & 2.10 & 0.92 \\
  CWVAE & \textbf{0.43} & \textbf{2.85} & \textbf{0.96} \\
  \hline
\end{tabular}
\caption{Human evaluation results on Event2Mind.}
\label{table:human_eva_e}
\end{table}

\begin{table}[]
  \setlength{\abovecaptionskip}{0.1cm}
  \setlength{\belowcaptionskip}{-0.3cm}
\footnotesize
\centering
\begin{tabular}{L{3cm}|L{1cm}C{1cm}C{1cm}}
  \hline
  Methods & Coherence & Diversity & Fluency  \\
  \hline
  RNN-based Seq2Seq & 0.21 & 2.66 & 0.78  \\
  Variational Seq2Seq & 0.22 & 2.70 & 0.90 \\
  VRNMT & 0.24 & 2.61 & 0.78 \\
  \hline
  \hline
  CWVAE-Unpretrained & 0.25 & 2.72 & 0.83 \\
  CWVAE & \textbf{0.32} & \textbf{3.03} & \textbf{0.90} \\
  \hline
\end{tabular}
\caption{Human evaluation results on Atomic.}
\label{table:human_eva_a}
\end{table}
\begin{table*}[h]
  \setlength{\abovecaptionskip}{0.1cm}
  \setlength{\belowcaptionskip}{-0.3cm}
\scriptsize
\centering
\begin{tabular}{p{1.6cm}p{0.8cm}p{3.8cm}p{3.8cm}p{3.5cm}}
  \hline
  \multirow{2}*{Base event} & \multirow{2}*{\tabincell{l}{Inference \\ dim.}} & \multicolumn{2}{c}{Generations}  & \multirow{2}*{Ground truth} \\ \cline{3-4}
  ~ & ~ & CWVAE & RNN-based Seq2Seq & ~ \\
  \hline
  \tabincell{l}{PersonX works \\ tirelessly } & \tabincell{l}{ xIntent} &\tabincell{l}{be productive and hardworking \\ finish his work soon \\ earn more money and accomplish goal
 \\ finish his work and make money \\ finish his work and accomplish goal \\ be productive and successful} & \tabincell{l}{get the job done \\ get the job done on time \\ get the job done in the way \\ get the job done for his life \\ make money for his own future \\ make money for his own money} & \tabincell{l}{be productive \\ finish the project as soon as possible \\ reach goal } \\
  \hline
\end{tabular}
\caption{An example of inferences made by CWVAE and RNN-based Seq2Seq model under inference dimension ``xIntent''.}
\label{table:case}
\end{table*}


To further evaluate the effectiveness of our proposed approach, we also conduct human evaluations, the results of which are shown in Table~\ref{table:human_eva_e} and Table~\ref{table:human_eva_a}. On both datasets, CWVAE-based methods achieve consistent better coherence, diversity and fluency performances. While comparing with CWVAE-Unpretrained, the pretrain procedure could improves the performance on coherence and fluency. The main reasons are twofold: first, the CWVAE has advantage in capturing the semantic distribution of targets; second, event background learned from the pretrain stage is helpful for the If-Then reasoning.

\subsection{Case Study}
Table~\ref{table:case} provides an example of model generations given the base event ``PersonX works tirelessly'' and the inference dimension ``xIntent''. The generations under CWVAE mainly contain four kinds of semantics: (1) be productive, (2) finish his work soon, (3) accomplish goal, (4) earn more money. While the semantics of generations using baseline RNN-based Seq2Seq model is relatively limited. Furthermore, the first three kinds of semantic overlap the three ground truth targets, and the fourth kind of semantic is in accordance with daily-life commonsense. Compared to RNN-based Seq2Seq model, our approach can increase the diversity and rationality of generations, meanwhile keep the accuracy.


\section{Related Work}
\subsection{Event-Centered Commonsense Reasoning}
Understanding events and constructing event-centered commonsense knowledge are crucial to many NLP applications, such as intention recognition \cite{goldwasser2016understanding} and dialog generation \cite{wen2017latent}. Recently a growing number of studies focus on event-centered commonsense reasoning, which mainly concentrates on two areas, script event prediction and story ending generation/choosing.


Script event prediction concerns with the temporal relationships between script events \cite{granroth2016happens}, which requires models to choose a correct subsequent triple-organized event among the candidates \cite{wang2017integrating}. Prior work mainly focused on modeling event pairs \cite{granroth2016happens}, event chains \cite{wang2017integrating} and event graph \cite{li2018constructing} to predict the subsequent event. Story ending generation focuses on generating plausible story endings \cite{mostafazadeh2016corpus}, which requires models to understand the story context, and keep generated endings logically consistent with it \cite{peng2017joint,guan2019story}. The above tasks mainly investigate the logical orders of events, whereas the If-Then reasoning task focuses on inferring the mental state of event participants.


\subsection{Variational AutoEncoder-Decoder Based Natural Language Generation}

VAE \cite{kingma2014auto} has been widely applied in various of text generation tasks, such as dialogue and machine translation. 

In dialogue generation, \citeauthor{zhao2017learning} (\citeyear{zhao2017learning}) adapts VAE with encoder-decoder framework to model the latent semantic distribution of answers, which can increase the diversity of generations. For the task of machine translation, \citeauthor{su2018variational} (\citeyear{su2018variational}) and \citeauthor{zhang2016variational} (\citeyear{zhang2016variational}) employ a latent variable to capture the semantic interaction between the source and target sentence, and regard the latent variable as a supplementation of attention mechanism. While \citeauthor{Wang2019Topic} (\citeyear{Wang2019Topic}) use the latent variable to model topic distributions in text generation. In this paper, we introduce an additional context-aware latent variable to effectively learn background knowledge and conduct If-Then reasoning on the guidance of it.

\section{Conclusion}

In this paper, we propose a novel context-aware VAE (CWVAE) framework with two training stages for If-Then commonsense reasoning. By introducing an additional context-aware latent variable, CWVAE is able to learn external background knowledge, and conduct If-Then reasoning under its guidance. In the pretrain stage, CWVAE learns event background knowledge, then in the finetune stage CWVAE adapts such knowledge to each inference dimension. Experimental results demonstrate that CWVAE outperforms baseline methods in both the accuracy and diversity of generations.

\section{Acknowledgments}
We thank the anonymous reviewers for their constructive comments, and gratefully acknowledge the support of the National Key Research and Development Program of China (SQ2018AAA010010), the National Key Research and Development Program of China (2018YFB1005103), the National Natural Science Foundation of China (NSFC) via Grant 61702137.

\bibliography{emnlp-ijcnlp-2019}
\bibliographystyle{acl_natbib}

\end{document}